\documentclass[letterpaper]{article} 
\usepackage{arxiv}  

\usepackage[hyphens]{url}  
\usepackage{graphicx} 
\usepackage{natbib}  
\usepackage{caption} 
\usepackage{algorithm}
\usepackage{algorithmic}

\usepackage{newfloat}
\usepackage{listings}
\DeclareCaptionStyle{ruled}{labelfont=normalfont,labelsep=colon,strut=off} 
\floatstyle{ruled}
\newfloat{listing}{tb}{lst}{}
\floatname{listing}{Listing}

\usepackage{booktabs}
\usepackage{amsmath}
\usepackage{amssymb}
\usepackage{makecell}
\usepackage[table]{xcolor}
\usepackage{multirow}

\author{
Hanze Jia\textsuperscript{\rm 1},
Chunshi Wang\textsuperscript{\rm 2},
Yuxiao Yang\textsuperscript{\rm 1},
Zhonghua Jiang\textsuperscript{\rm 1},\\
Yawei Luo\textsuperscript{\rm 2},
Shuainan Ye\textsuperscript{\rm 1},
Tan Tang\textsuperscript{\rm 3}
}

\affiliations{
\textsuperscript{\rm 1}State Key Lab of CAD\&CG, Zhejiang University
\textsuperscript{\rm 2}School of Software Technology, Zhejiang University
\textsuperscript{\rm 3}Laboratory of Art and Archaeology Image, Zhejiang University
\{hzjia,yuxiaoyang,jiangzhonghua,snye\}@zju.edu.cn
\{chunshiwang,yaweiluo\}@zju.edu.cn,
tangtan@zju.edu.cn
}

\title{SARe: Structure-Aware Generative 3D Fragment Reassembly}

\begin{document}

\maketitle

\begin{abstract}
3D fragment reassembly estimates the rigid pose of each fragment to recover a complete object from unordered point clouds or meshes.
The task becomes increasingly challenging as the fragment count grows, since irregular fragments provide weak semantic cues and admit rapidly increasing numbers of plausible contact relations and global configurations.
We propose Structure-Aware Reassembly (SARe), a generative framework that integrates query-aligned local geometry and task-native structural supervision into point-flow assembly.
SARe-Gen conditions each transported surface query on a local latent and jointly supervises intermediate flow tokens with query-level fracture-region and fragment-level contact targets.
Because these heads are optimized together with flow matching, structural supervision directly shapes the representations that drive coordinate transport, without requiring additional reassembly-specific pretraining or a separate teacher-alignment stage.
At inference time, SARe-Refine geometrically verifies predicted relations and uses reliable local subassemblies to guide a second sampling pass, reinforcing consistent regions while resampling uncertain fragments.
We evaluate SARe across three settings, including synthetic fractures, simulated fractures from scanned real objects, and scans of physically fractured objects.
The results demonstrate state-of-the-art performance, with higher part accuracy and more graceful degradation in
challenging many-fragment settings.
\end{abstract}

\section{Introduction}
\label{sec:intro}

3D fragment reassembly aims to recover the rigid pose of each fragment from unordered point clouds or meshes, with applications in digital heritage restoration, robotic assembly, and accident reconstruction ~\cite{accident1,robotic1,heritage1,heritage2-RePAIR,robotic2}.
Although the output is a set of fragment poses, their compatibility is supported by relational contact cues: which fragment pairs share a mating fracture contact and where their corresponding fracture surfaces should meet.
The problem is challenging because the complete shape is unknown and fragments are irregular with few semantic cues~\cite{Tsesmelis2024RePAIR,Lu2025SurveyAssembly}.
As the fragment count grows, the number of plausible relations and global configurations increases rapidly, making local errors more likely to accumulate.

\begin{figure}[t]
  \centering
  \includegraphics[width=0.99\columnwidth]{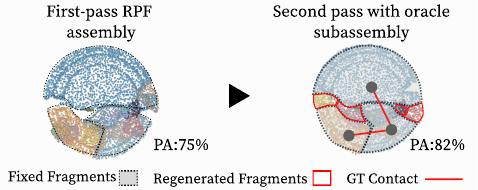}
  \caption{
    Representative oracle partial-structure diagnostic on RPF.
Preserving a local subassembly specified by ground-truth contact
edges while regenerating the remaining fragments improves PA from
75\% to 82\% in this example.
  }
  \label{fig:intro}
\end{figure}

Early methods relied on explicit geometric matching and global optimization, yielding reasonable results under regular fracture patterns or with a small number of fragments~\cite{Funkhouser2011,Structure_From_Sherds}.
Large-scale fracture datasets~\cite{BreakingBad} subsequently enabled learning-based methods that infer cross-fragment interactions and fragment poses directly from geometry~\cite{Wu_2023_SE3,Lu2023Jigsaw,Qin2022Geometric,Huang2020GPA}.
Recent generative approaches either denoise per-fragment poses~\cite{GARF,FragmentDiff,JigsawPP,PuzzleFusionPP}, or transport points into a shared frame and recover poses afterward~\cite{RPF}; the latter avoids direct rotation generation and preserves input--output point correspondence, naturally supporting point-aligned geometric conditioning.
Beyond the choice of generated variables, a central question is how geometric priors are introduced into point-flow assembly models.
RPF~\cite{RPF} uses features from a task-specifically pretrained overlap-aware encoder, whereas TORA~\cite{lee2026tora} distills point-wise descriptors and pairwise feature similarities from a frozen 3D teacher into intermediate flow representations.
Meanwhile, modern 3D generative models represent shapes using spatially grounded local latents associated with explicit query locations~\cite{Zhang2023_3DShape2VecSet,zhang2025lagem,Zhao2025Hunyuan3D2}.
This suggests a more direct formulation for fragment reassembly: rather than introducing a separate teacher-alignment stage, local geometry can be attached directly to each transported surface query, providing query-aligned neighborhood information for coordinate generation, fracture-region localization, and rigid-pose recovery.
We further observe that an imperfect first-pass assembly may already contain locally correct fragment relations.
As an oracle diagnostic on RPF, we use a small set of ground-truth contact edges to specify a local subassembly, keep its fragments fixed, and regenerate the remaining fragments during a second inference pass, as shown in Figure~\ref{fig:intro}.
This result suggests that a reliable local subassembly can serve as an actionable inference condition for correcting the remaining uncertain fragments.
It also inspires us to feed reliable structural information back into generation as an inference-time condition, thereby closing the assembly-generation loop.

Motivated by these observations, we propose \textbf{Structure-Aware Reassembly (SARe)}, a generative framework that couples query-aligned point transport with instance-specific contact structure.
Here, ``structure'' refers to two complementary aspects of an assembly: which surface regions participate in mating fracture contacts and which fragment pairs share such contacts.
Rather than aligning intermediate flow features to an external teacher space, \textbf{SARe-Gen} conditions each transported surface query on a local ShapeVAE latent~\cite{Zhao2025Hunyuan3D2}, providing direct access to pretrained local geometry throughout sampling.
Inspired by the broader principle of intermediate representation supervision~\cite{yu2025repa}, SARe-Gen supervises intermediate flow tokens through task-native heads that predict query-level fracture-region probabilities and a fragment-level contact graph.
Jointly optimized with the flow-matching objective, these heads shape the same backbone tokens that drive coordinate transport, without requiring a separate structural pretraining stage.
The resulting fracture and contact predictions are retained as instance-specific hypotheses for downstream verification and correction.
At inference time, \textbf{SARe-Refine} filters candidate contact edges through geometric-consistency checks to retain reliable local subassemblies.
These verified subassemblies then guide a second sampling pass that reinforces their first-pass configurations while resampling the remaining uncertain fragments.
Finally, we evaluate SARe across synthetic fractures, simulated fractures of scanned objects, and real fractured scans.
SARe-Gen achieves state-of-the-art across all three settings, with pronounced gains in many-fragment cases, while SARe-Refine further improves difficult assemblies through inference-time structural correction.

In summary, our key contributions are:
(1) We introduce \textbf{SARe}, consisting of \textbf{SARe-Gen} for query-aligned assembly generation with task-native fracture and contact supervision, and \textbf{SARe-Refine} for geometry-verified, subassembly-guided resampling.
(2) We build and release a scan-derived simulated-fracture benchmark containing ${\sim}55$K samples generated from real-world scanned objects, together with structural supervision signals.
(3) Experiments across three data settings demonstrate state-of-the-art results with more graceful degradation at high fragment counts.

\section{Related work}
\label{related_work}
\subsection{3D Fracture Reassembly.}
Unlike semantic part assembly, fracture reassembly estimates the rigid poses of irregular fragments with weak semantics, relying heavily on fracture-surface compatibility and global inter-fragment consistency~\cite{Lu2025SurveyAssembly,Chen2022NeuralShape,Li2024GPAT,heritage2-RePAIR}.
Classical methods matched hand-crafted geometric features and optimized global configurations, but struggled with ambiguity and combinatorial complexity as the fragment count increased~\cite{Funkhouser2011,Structure_From_Sherds,Huang2006Reassembling,
Xu2015RobustSurface,Castaneda2011GlobalConsistency}.
Large-scale benchmarks~\cite{BreakingBad} subsequently enabled learning-based methods that infer fragment relations, poses, or complete-shape priors directly from geometry~\cite{Wu_2023_SE3,Qin2022Geometric,Lu2023Jigsaw,JigsawPP}.
Recent generative approaches denoise fragment poses~\cite{Scarpellini2024DiffAssemble,PuzzleFusionPP,GARF}, or
transport points into an assembled frame and recover poses afterward~\cite{RPF}.
FragmentDiff~\cite{FragmentDiff} additionally predicts fragment adjacency as an auxiliary task.
Among point-flow methods, RPF relies on a task-specifically pretrained overlap-aware encoder, while TORA~\cite{lee2026tora} aligns intermediate representations with point-wise and pairwise signals from a frozen teacher.
In contrast, SARe directly shapes flow representations with task-native fracture and contact supervision, then geometrically verifies the predicted structure and feeds it back for subassembly-guided resampling, without extra reassembly-specific pretraining or teacher alignment.

\subsection{3D Diffusion and Flow Models.}
Modern 3D generative models commonly learn compact shape representations with an autoencoder and perform diffusion or flow matching in latent or token spaces~\cite{Zeng2022LION,Zhang2023_3DShape2VecSet,TRELLIS}.
Representative designs include sparse volumetric latents~\cite{TRELLIS,Li2025Sparc3D,Ren2024Xcube} and set-based local latents~\cite{Zhang2023_3DShape2VecSet,zhang2025lagem,Chen2025Dora}.
Recent work further demonstrates the compositionality of tokenized representations, where subsets of local tokens can be decoded into the corresponding geometric regions~\cite{Chen2025AutoPartGen,Zhao2025Assembler}.
Motivated by these advances, we use query-aligned local tokens from Hunyuan3D-ShapeVAE~\cite{Zhao2025Hunyuan3D2} as spatially grounded fragment representations that encode fine-grained local surface geometry.

\subsection{Representation Supervision in Diffusion Models.}
Diffusion models can leverage pretrained representations as semantic conditions or intermediate alignment targets~\cite{Pernias2024Wuerstchen,Li2024ReturnOfUnconditionalGeneration,yu2025repa}.
A complementary line jointly predicts task-specific auxiliary structures alongside the denoising objective, including masks, segmentation, and dense task representations~\cite{Guo_2023_CVPR,Xu2025TopoDiffuser,Ye2024DiffusionMTL,Yang2025TaskDiffusion}.
Rather than aligning to an external feature space, SARe supervises intermediate flow tokens with task-native fracture-region and contact-relation targets, jointly optimized with flow matching to shape the generative backbone.

\section{Reassembly Problem Formulation}

Given a fractured object consisting of $K$ fragments, we represent each fragment by a point set (or surface samples from a mesh) $\{\mathcal{P}_i\}_{i=1}^{K}$, 
where
$\mathcal{P}_i=\{\mathbf{x}_{i,n}\in\mathbb{R}^3\}_{n=1}^{N_i}$
is defined in the local coordinate system of fragment $i$.
The goal of 3D fragment reassembly is to recover a rigid transformation
$\mathcal{T}_i=(R_i,\mathbf{t}_i)\in SE(3)$ for each fragment, where
$R_i\in SO(3)$ and $\mathbf{t}_i\in\mathbb{R}^3$,
so that the transformed fragment becomes
\begin{equation}
\mathcal{P}'_i \;=\; \mathcal{T}_i(\mathcal{P}_i)\;=\;\{\,R_i\mathbf{x}+\mathbf{t}_i \mid \mathbf{x}\in\mathcal{P}_i\,\},
\label{eq:rigid_transform}
\end{equation}
and the assembled shape is $\mathcal{X}'=\bigcup_{i=1}^{K}\mathcal{P}'_i$ in a common object coordinate system.

Following RPF's Euclidean point-space formulation~\cite{RPF} and inspired by point-based 3D generative models~\cite{Lan2025GaussianAnything,Zhao2025Assembler}, we model reassembly as conditional generation of assembled query coordinates rather than direct pose generation.
Specifically, let $\mathcal{Q}_i=\{\mathbf{q}_{i,m}\}_{m=1}^{M_i}\subset\mathcal{P}_i$ denote surface queries sampled from fragment $i$.
The model predicts their corresponding assembled coordinates $\mathcal{Q}'_i=\{\mathbf{q}'_{i,m}\}_{m=1}^{M_i}$ in the shared object frame, from which $\mathcal{T}_i$ is recovered by rigid alignment between $\mathcal{Q}_i$ and $\mathcal{Q}'_i$.
Although the target output is a set of fragment poses, each ground-truth assembly also induces task-native structural variables describing \emph{which} fragment pairs share a mating contact and \emph{where} the contacts occur.
We represent the fragment-contact graph by a binary contact matrix $A\in\{0,1\}^{K\times K}$, where $A_{ij}=1$ indicates that fragments $i$ and $j$ share a mating fracture contact in the ground-truth assembly.
For each query $\mathbf{q}_{i,m}$, a fracture-region label $f_{i,m}\in\{0,1\}$ indicates whether it lies on a mating fracture region, and we denote $F_i=\{f_{i,m}\}_{m=1}^{M_i}$.
SARe jointly predicts the assembled query coordinates $\{\mathcal{Q}'_i\}$, a contact matrix $\hat A\in[0,1]^{K\times K}$, and query-level fracture-region probabilities $\{\hat F_i\}$, where $\hat F_i=\{\hat f_{i,m}\}_{m=1}^{M_i}$.
The ground-truth variables $A$ and $\{F_i\}$ provide task-native supervision to the generative backbone, while their predictions $\hat A$ and $\{\hat F_i\}$ are retained as instance-specific hypotheses for inference-time verification.

\section{Method}

\begin{figure*}[t]
  \centering
  \includegraphics[width=0.99\textwidth]{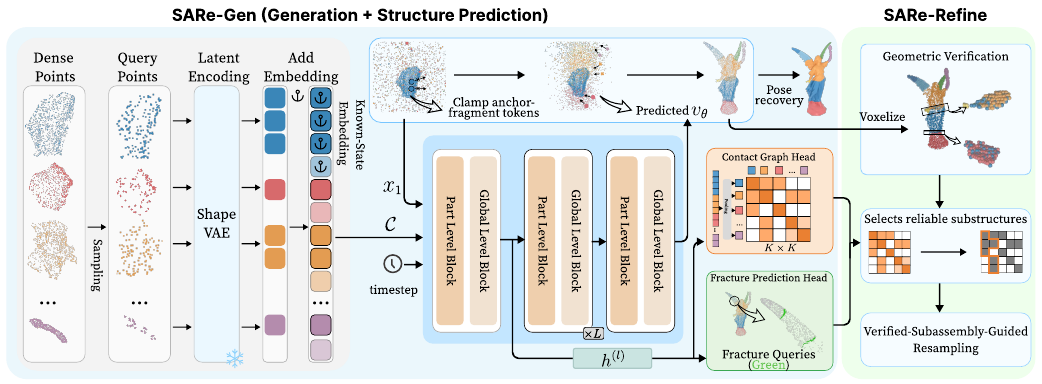}
  \caption{\textbf{Overview of SARe.}
  SARe-Gen transports query-aligned points into the assembled frame while jointly predicting query-level fracture-region probabilities and a fragment-level contact graph.
  SARe-Refine geometrically verifies the predicted relations and uses the retained local subassemblies to guide second-pass resampling.}
  \label{fig:pipeline}
\end{figure*}

We first construct query-aligned geometric conditions, then introduce SARe-Gen for joint assembly generation and structural prediction,  and finally present SARe-Refine, which verifies predicted relations and uses reliable subassemblies to guide second-pass resampling.

\subsection{Query-Point Conditioning}
To instantiate the query-level formulation in Sec.~3, we represent each fragment by an area-adaptive set of surface queries under a fixed object-level sampling budget.
Concretely, for fragment $i$ we first sample a dense surface point set via importance sampling, and then select $M_i$ query points using Farthest Point Sampling, 
yielding $\mathcal{Q}_i=\{q_{i,m}\}_{m=1}^{M_i}$ with $\mathcal{Q}_i\subset\mathcal{P}_i$.
The allocation satisfies $M=\sum_{i=1}^{K}M_i$, with $M_i$ proportional to the fragment surface area.
Here, $M$ controls the sampling density rather than the architectural dimension, allowing the backbone to process variable-length query sets.
For each query point, we keep its input-frame coordinate and normal $(q_{i,m}^{in},n_{i,m}^{in})$ and apply a positional encoding $\gamma(\cdot)$~\cite{Tancik2020FourierFeatures,Mildenhall2020NeRF}.
To capture local fracture geometry beyond sparse coordinates, we leverage a pretrained and frozen shape encoder ShapeVAE~\cite{Zhao2025Hunyuan3D2} to perform query-based tokenization.
Specifically, the encoded query locations serve as queries and the dense surface features as keys and values in cross-attention, yielding a geometric latent $\mathbf{z}_{i,m}$ aligned with each $\mathbf{q}^{in}_{i,m}$.
To preserve fragment membership after concatenating all queries, we add a fragment-ID embedding $\mathbf{e}^{part}_i$ to the queries of fragment $i$.
We additionally use a binary known-fragment indicator $s_i\in\{0,1\}$, whose embedding $\mathbf{e}^{known}(s_i)$ specifies whether fragment $i$ is supplied as a known condition.
The conditioning token is
\begin{equation}
\mathbf{c}_{i,m}
=
\phi\!\left(
\left[
\mathbf{z}_{i,m};
\gamma(\mathbf{q}^{in}_{i,m});
\gamma(\mathbf{n}^{in}_{i,m})
\right]\right)
+
\mathbf{e}^{part}_i
+
\mathbf{e}^{known}(s_i),
\label{eq:query_condition}
\end{equation}
where $\phi(\cdot)$ denotes a learnable linear projection.
The known-fragment indicator generalizes the conventional single-anchor condition to arbitrary known subassemblies.
Following the common practice of fixing a reference fragment~\cite{GARF,PuzzleFusionPP}, we select the fragment with the largest $M_i$ as the anchor.
During first-pass generation, only the anchor is marked as known ($s_i=1$) to fix the global gauge.
During second-pass refinement, the indicator additionally identifies the verified fragments used to guide second-pass resampling, while the remaining fragments have $s_i=0$.
Thus, the same conditioning interface supports both single-anchor generation and subassembly-guided refinement.
Aggregating all query tokens yields the point-aligned condition set $\mathcal{C}=\{\mathbf{c}_{i,m}\}$, which conditions the subsequent reassembly dynamics.

\subsection{SARe-Gen: Structure-Aware Rectified Flow for Reassembly}
Given the point-aligned conditioning tokens $\mathcal{C}$ from the previous subsection, we model reassembly as conditional rectified flow to generate the assembled query points, while jointly predicting structural variables.

\noindent{\textbf{Rectified Flow}.}
Following rectified flow with linear interpolation and conditional flow matching~\cite{RPF,Liu2023RectifiedFlow,Liu2022RectifiedFlowOT}, let $\mathcal{Q}'=\bigcup_{i=1}^{K}\mathcal{Q}'_i$
denote the ground-truth assembled query set during training, flattened as $x_0\in\mathbb{R}^{M\times3}$.
We sample a noise endpoint $x_1\sim\mathcal{N}(0,I)$ and $t\sim\mathcal{U}[0,1]$ to construct
\begin{equation}
x_t=(1-t)x_0+t x_1,\qquad v_t=\frac{dx_t}{dt}=x_1-x_0.
\end{equation}
The conditional velocity field $v_\theta(x_t,t\mid\mathcal{C})$ is optimized by
\begin{equation}
\mathcal{L}_{\mathrm{rf}}=\mathbb{E}\Big[\big\|v_\theta(x_t,t\mid\mathcal{C})-v_t\big\|_F^2\Big].
\end{equation}
At inference, we initialize from $x_1$ and integrate the learned field from $t=1$ to $t=0$.
We implement $v_\theta$ with a DiT-style transformer that processes the noisy coordinates $x_t$, injects their aligned conditions $\mathcal{C}$, and predicts one velocity vector for each query. 
Token ordering is preserved throughout the flow, so each generated coordinate $\mathbf{q}'_{i,m}$ remains associated with its input query $\mathbf{q}^{in}_{i,m}$.

\noindent{\textbf{Joint structural prediction.}}
To inject task-native structural supervision into the generative backbone, we attach two lightweight prediction heads to the intermediate flow tokens $\mathbf{h}^{(\ell_s)}\in\mathbb{R}^{M\times D}$, where $\ell_s$ denotes the attachment layer.
Inspired by intermediate representation supervision~\cite{yu2025repa} and auxiliary multi-task diffusion~\cite{Xu2025TopoDiffuser,Ye2024DiffusionMTL,Yang2025TaskDiffusion}, the heads are optimized jointly with the flow-matching objective, so their supervision directly shapes the same intermediate representations that drive coordinate transport, without requiring a separate task-specific structural pretraining stage.
A token-level MLP predicts fracture-region probabilities $\hat f_{i,m}\in[0,1]$.
For contact prediction, we pool the intermediate tokens within each fragment to obtain fragment-level features and score fragment pairs, yielding a predicted contact matrix $\hat A\in[0,1]^{K\times K}$.
The fracture-region head and contact-graph head are trained with binary cross-entropy losses $\mathcal{L}_{F}$ and $\mathcal{L}_{C}$, respectively, using the fracture-region labels $\{\mathbf{F}_i\}$ and contact matrix $\mathbf{A}$ obtained during dataset preprocessing.

\noindent{\textbf{Overall objective and anchor constraint.}}
We jointly optimize
\begin{equation}
\mathcal{L}
=
\mathcal{L}_{\mathrm{rf}}
+
\lambda_F\mathcal{L}_F
+
\lambda_C\mathcal{L}_C.
\end{equation}
During first-pass generation, only the reference anchor is hard-clamped: its query coordinates are kept at their supplied positions throughout sampling to fix the global reference frame.

\noindent{\textbf{Pose recovery.}}
We recover the rigid transform $\mathcal{T}_i=(R_i,\mathbf{t}_i)$ for each fragment by aligning its input-frame query points
$\{q^{in}_{i,m}\}_{m=1}^{M_i}$ with the corresponding assembled query points $\{q'_{i,m}\}_{m=1}^{M_i}$.

\subsection{SARe-Refine: Inference-Time Structure-Guided Refinement}

\begin{figure}[t]
  \centering
  \includegraphics[width=0.99\columnwidth]{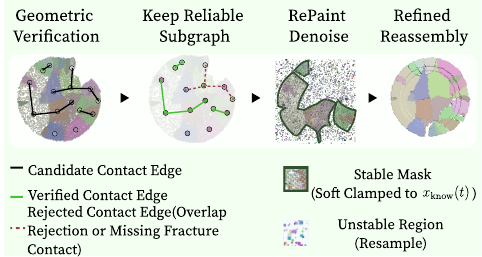}
  \caption{\textbf{SARe-Refine pipeline.}
Predicted contact edges are filtered to identify reliable substructures, after which uncertain regions are resampled in a RePaint-style
  process conditioned on a stable mask.}
  \label{fig:refine}
\end{figure}

SARe-Refine is an optional inference-time procedure for correcting imperfect first-pass assemblies, particularly in many-fragment cases.
As illustrated in Fig.~\ref{fig:refine}, it geometrically verifies the structural predictions of SARe-Gen and retains one or more reliable local subassemblies as structural conditions for a second resampling pass.
The reference anchor remains hard-clamped, whereas the other verified fragments are softly reinforced rather than fixed.

We first run SARe-Gen to obtain initial assembled query coordinates $x^{(0)}\in\mathbb{R}^{M\times3}$, together with the query-level fracture-region probabilities $\hat f_{i,m}$ and fragment-level contact score matrix $\hat A$ produced at the final denoising step $t=0$.
Because directly thresholding $\hat{\mathbf{A}}$ may introduce spurious contact relations, we first construct a candidate contact edge set $\mathcal{E}_{\mathrm{cand}}$ and apply geometric verification to each candidate edge.
For each candidate contact edge $(i,j)\in\mathcal{E}_{\mathrm{cand}}$, we voxelize the fragments under their predicted poses to obtain the occupied voxel set $S_i^v$ and $S_j^v$, and transform the predicted fracture-region queries into the assembled frame to obtain fracture-region voxel sets $F_i^v$ and $F_j^v$.
Pairs with excessive interpenetration, measured by $r_{ij}=|S_i^v\cap S_j^v|/\min(|S_i^v|,|S_j^v|)>\tau_o$, are discarded.
Among the remaining pairs, we retain only those whose fracture regions exhibit sufficient mutual overlap within a small voxel tolerance.
The verified contact edges $\mathcal{E}_{\mathrm{keep}}$ define a verified fragment-contact
subgraph whose connected components above a minimum size form reliable local subassemblies, and the indices of their query tokens are collected in a stable mask $m_{\mathrm{st}}$.

In the second sampling pass, the first-pass coordinates of tokens in $m_{\mathrm{st}}$ define the reference state $x_{\mathrm{ref}}=x^{(0)}[m_{\mathrm{st}}]$.
Under the straight-line coupling of rectified flow, the noise-consistent state for this region at timestep $t$ is $x_{\mathrm{known}}(t)=(1-t)x_{\mathrm{ref}}+t\epsilon$, where $\epsilon$ is the noise endpoint. 
Inspired by RePaint-style inpainting~\cite{Lugmayr2022RePaint,Go2025SplatFlow}, we reinforce the stable region before each model evaluation and after each numerical step:
\begin{equation}
x_t[m_{\mathrm{st}}]\leftarrow(1-\alpha)\,x_t[m_{\mathrm{st}}]+\alpha\,x_{\mathrm{known}}(t),
\end{equation}
where $\alpha\in[0,1]$ controls the injection strength.
Only the reference anchor remains hard-clamped; the other stable tokens continue to evolve under the flow and are repeatedly reinforced throughout the trajectory.
All hyperparameters are fixed and reported in the supplementary material.

\section{Experiments}


\begin{table*}[t]
\centering
\setlength{\tabcolsep}{2.0pt}
\renewcommand{\arraystretch}{1.08}
\small
\begin{tabular}{l*{4}{cccc}}
\toprule
& \multicolumn{4}{c}{Everyday, $K=2$--$20$}
& \multicolumn{4}{c}{Everyday, $K=21$--$50$}
& \multicolumn{4}{c}{Artifact, $K=2$--$20$}
& \multicolumn{4}{c}{Artifact, $K=21$--$50$} \\
\cmidrule(lr){2-5}
\cmidrule(lr){6-9}
\cmidrule(lr){10-13}
\cmidrule(lr){14-17}
Method
& \makecell{PA$\uparrow$\\(\%)}
& \makecell{CD$\downarrow$\\$\times10^{-3}$}
& \makecell{RE$\downarrow$\\(deg.)}
& \makecell{TE$\downarrow$\\$\times10^{-2}$}
& \makecell{PA$\uparrow$\\(\%)}
& \makecell{CD$\downarrow$\\$\times10^{-3}$}
& \makecell{RE$\downarrow$\\(deg.)}
& \makecell{TE$\downarrow$\\$\times10^{-2}$}
& \makecell{PA$\uparrow$\\(\%)}
& \makecell{CD$\downarrow$\\$\times10^{-3}$}
& \makecell{RE$\downarrow$\\(deg.)}
& \makecell{TE$\downarrow$\\$\times10^{-2}$}
& \makecell{PA$\uparrow$\\(\%)}
& \makecell{CD$\downarrow$\\$\times10^{-3}$}
& \makecell{RE$\downarrow$\\(deg.)}
& \makecell{TE$\downarrow$\\$\times10^{-2}$} \\
\midrule

Jigsaw
& 57.30 & 13.30 & 42.30 & 10.70
& --    & --    & --    & --
& 45.60 & 14.30 & 52.40 & 22.20
& --    & --    & --    & -- \\

PF++
& 72.40 & 3.65 & 39.20 & 8.46
& --    & --   & --    & --
& 53.53 & 7.63 & 51.26 & 14.37
& --    & --   & --    & -- \\
\midrule
GARF
& 93.22 & 1.10 & 20.01 & 4.17
& 51.80 & 5.99 & 75.52 & 31.97
& 89.22 & 2.79 & 29.87 & 4.09
& 37.23 & 6.30 & 68.84 & 29.41 \\

RPF
& 93.39 & 0.78 & 20.60 & 3.90
& 58.51 & 4.70 & 75.18 & 21.56
& 92.76 & 1.60 & 25.60 & 4.06
& 57.16 & 5.92 & 75.64 & 22.28 \\

TORA-CKA
& 94.92 & 0.58 & 15.42 & 3.84
& 71.74 & \textbf{1.58} & 65.33 & 15.23
& 94.13 & 0.86 & 21.45 & 5.46
& 68.39 & 3.36 & 64.73 & 15.82 \\
\midrule
SARe-Gen
& \textbf{96.75} & \textbf{0.33} & \textbf{10.11} & \textbf{1.24}
& \textbf{83.05} & 2.26 & \textbf{40.05} & \textbf{8.27}
& \textbf{96.10} & \textbf{0.62} & \textbf{8.19} & \textbf{1.30}
& \textbf{82.65} & \textbf{0.59} & \textbf{33.52} & \textbf{7.48} \\

\bottomrule
\end{tabular}

\caption{
Quantitative comparison on the non-volume-constrained Breaking Bad benchmark.
Results are reported on the Everyday and Artifact subsets under different
fragment-count ranges.
}
\label{tab:main_nonvol}
\end{table*}

\subsection{Experimental Setting}
\noindent{\textbf{Implementation Details.}}
We use a frozen ShapeVAE~\cite{Zhao2025Hunyuan3D2} as a geometric encoder and allocate a total of $M=5120$ surface queries per object.
The generator follows the DiT-style architecture of RPF~\cite{Peebles2022DiT,RPF}.
We attach the fracture-region and contact prediction heads to the intermediate representation $h^{(\ell_s)}$ at layer $\ell_s=4$.
The model is trained for 200 epochs using AdamW with loss weights $\lambda_F=\lambda_C=0.01$.
The learning rate is fixed at $1\times10^{-4}$ for the first 100 epochs and then cosine-annealed to $1\times10^{-5}$ over the remaining 100 epochs.
Training is performed on four NVIDIA A100 GPUs with a global batch size of 32.
At inference, we integrate the rectified-flow trajectory using 50 Euler steps.
For SARe-Refine, the blending strength is set to $\alpha=0.5$.
Additional implementation details are provided in the supplementary material.

\noindent{\textbf{Datasets.}}
We evaluate SARe across synthetic CAD fractures, scan-derived simulated fractures, and real-world broken objects.
Our main benchmark is the original non-volume-constrained Breaking Bad dataset~\cite{BreakingBad}, including the Everyday and Artifact subsets with $K=2$--$50$ fragments.
Approximately 15\% of its samples contain more than 20 fragments, compared with less than 1\% in the commonly used volume-constrained variant, whose results are reported in the supplementary material.
To evaluate transfer to scanned geometry, we construct OmniObject3D-Fracture from 1,223 scanned objects in OmniObject3D~\cite{wu2023omniobject3d} using the Breaking Good fracture pipeline~\cite{breakinggood}, yielding 54,966 fracture instances.
The splits are constructed at the source-object level, ensuring that fracture instances generated from the same scanned object do not appear across different data splits.
We further evaluate on FRACTURA~\cite{GARF} and Fantastic Breaks~\cite{lamb2023fantastic}, which contain real fractures and scans of physically fractured objects.
All models are trained across all object categories in each benchmark's training split.

\noindent{\textbf{Evaluation Metrics.}}
Following GARF, we report the rotation error $\mathrm{RMSE}(R)$ and the translation error $\mathrm{RMSE}(T)$ as the root mean squared errors between the predicted and ground-truth fragment poses.
Part Accuracy (PA) is the fraction of fragments whose Chamfer distance between the predicted and ground-truth transformed point sets is below 0.01.
Finally, we report the object-level Chamfer Distance (CD) between the assembled shape and the ground-truth object.
For compactness, RE and TE denote RMSE(R) and RMSE(T), respectively.
CD is reported in $\times10^{-3}$, rotation error in degrees, and translation error in $\times10^{-2}$.

\noindent{\textbf{Competing Methods.}}
We compare with Jigsaw~\cite{Lu2023Jigsaw}, PuzzleFusion++~\cite{PuzzleFusionPP}, GARF~\cite{GARF}, RPF~\cite{RPF}, and the strongest CKA variant of TORA ~\cite{lee2026tora}.
GARF, RPF, and TORA-CKA are retrained on the same non-volume-constrained Breaking Bad splits and evaluated under our unified protocol.
Since the official GARF implementation supports at most 20 fragments, we extend its input handling to $K\leq50$.
Results for Jigsaw and PuzzleFusion++ are included as reference results under their available $K\leq20$ settings.

\begin{table}[t]
\centering
\small
\setlength{\tabcolsep}{4.5pt}
\renewcommand{\arraystretch}{1.08}

\begin{tabular}{lcccc}
\toprule
Method
& \makecell{PA$\uparrow$}
& \makecell{CD$\downarrow$}
& \makecell{RE$\downarrow$}
& \makecell{TE$\downarrow$} \\
\midrule
GARF
& 80.12 & 1.32 & 22.54 & 5.37 \\

RPF
& 82.32 & 1.35 & 20.06 & 4.33 \\

TORA-CKA
& \underline{93.75}
& \underline{0.47}
& \underline{14.78}
& \underline{1.94} \\
\midrule
SARe-Gen
& \textbf{97.68}
& \textbf{0.40}
& \textbf{9.92}
& \textbf{1.06} \\
\bottomrule
\end{tabular}
\caption{
Quantitative results on scan-derived simulated fractures.
OmniObject3D-Fracture is generated by applying simulated fracture processes
to real-world scanned objects.
}
\label{tab:omniobject}
\end{table}

\subsection{Quantitative Evaluations}

\noindent{\textbf{Multi-Fragment Reassembly on Synthetic and Scan-Derived Objects.}}
We first evaluate SARe-Gen on two complementary settings: simulated fractures of synthetic CAD objects from Breaking Bad and simulated fractures of real-world scanned objects from OmniObject3D-Fracture.
Table~\ref{tab:main_nonvol} reports results on the non-volume-constrained Breaking Bad benchmark, separately for the Everyday and Artifact subsets under $K=2$--$20$ and $K=21$--$50$.
Table~\ref{tab:omniobject} further evaluates assembly performance on scan-derived geometry from OmniObject3D-Fracture.
On Breaking Bad, SARe-Gen consistently achieves higher PA and lower rotation and translation errors than all competing methods across both subsets and fragment-count ranges.
The gains over the strongest point-flow baseline, TORA-CKA, are moderate for $K=2$--$20$ but increase substantially for $K=21$--$50$, indicating more graceful degradation as the number of fragments grows.
The same trend holds on OmniObject3D-Fracture, where SARe-Gen improves PA from 93.75\% to 97.68\% over TORA-CKA while also reducing Chamfer distance and pose errors.
These results show that the advantages of SARe extend beyond synthetic CAD geometry to simulated fractures of real-world scanned objects.

\noindent{\textbf{Zero-Shot Transfer to Unseen Fracture Datasets.}}
\begin{table}[t]
\centering
\small
\setlength{\tabcolsep}{2.5pt}
\renewcommand{\arraystretch}{1.05}

\resizebox{\columnwidth}{!}{
\begin{tabular}{lcccccccc}
\toprule
& \multicolumn{4}{c}{FRACTURA}
& \multicolumn{4}{c}{Fantastic Breaks} \\
\cmidrule(lr){2-5}
\cmidrule(lr){6-9}
Method
& PA$\uparrow$
& CD$\downarrow$
& RE$\downarrow$
& TE$\downarrow$
& PA$\uparrow$
& CD$\downarrow$
& RE$\downarrow$
& TE$\downarrow$ \\
\midrule
GARF
& 44.26 & 6.01 & 41.07 & 26.91
& 88.31 & 2.12 & 8.20 & 2.10 \\

RPF
& 68.11 & 3.64 & 51.15 & 11.23
& 96.90 & 2.53 & 6.32 & 2.18 \\

TORA-CKA
& \underline{76.35}
& \underline{1.89}
& \underline{36.40}
& \underline{7.72}
& \underline{97.28}
& \underline{0.26}
& \textbf{3.03}
& \underline{0.80} \\

\midrule
SARe-Gen
& \textbf{81.44}
& \textbf{0.59}
& \textbf{19.23}
& \textbf{4.14}
& \textbf{98.53}
& \textbf{0.23}
& 7.78
& \textbf{0.35} \\
\bottomrule
\end{tabular}
}
\caption{
Zero-shot transfer to unseen fracture datasets.
All methods are evaluated without fine-tuning.
}
\label{tab:zero_shot}
\end{table}
To evaluate robustness under domain shift, we train all models on the non-volume-constrained Breaking Bad Everyday subset and evaluate them without fine-tuning on two unseen datasets: FRACTURA, which contains mixed synthetic and real fractures, and Fantastic Breaks, which contains scans of physically broken real-world objects.
As shown in Table~\ref{tab:zero_shot}, SARe-Gen substantially improves zero-shot performance on FRACTURA, increasing PA from 76.35\% to 81.44\% over TORA-CKA while consistently reducing Chamfer distance and pose errors.
On Fantastic Breaks, SARe-Gen achieves the best PA, Chamfer distance, and translation error, whereas TORA-CKA obtains the lowest rotation error.
Overall, these results demonstrate that SARe transfers effectively from synthetic CAD fractures to unseen mixed-fracture and physically broken-object domains.

\noindent{\textbf{SARe-Refine on Difficult Cases.}}
To evaluate refinement where correction is most needed, we retrospectively pool the Everyday and Artifact test samples whose first-pass SARe-Gen PA is below 95\%; this threshold is used only for evaluation and is not accessed by SARe-Refine.
As shown in Table~\ref{tab:refine_hard}, SARe-Refine increases the mean PA from 81.34\% to 84.89\% (+3.55 percentage points), while reducing Chamfer distance from 1.85 to 1.72, rotation error from 34.35$^\circ$ to 32.51$^\circ$, and translation error from 6.83 to 6.19.
Although 75\% of test instances already achieve first-pass PA $\geq95\%$, SARe-Refine improves the full-set mean PA from 94.70\% to 95.34\%, with gains on 10.30\% of all instances. 
Thus, its benefit is not restricted to the retrospectively selected difficult subset.
These consistent improvements show that the verified-subassembly-guided second pass effectively corrects imperfect first-pass assemblies.

\begin{table}[t]
\centering
\small
\setlength{\tabcolsep}{5pt}
\renewcommand{\arraystretch}{1.08}

\begin{tabular}{lcccc}
\toprule
Method
& \makecell{PA$\uparrow$}
& \makecell{CD$\downarrow$}
& \makecell{RE$\downarrow$}
& \makecell{TE$\downarrow$} \\
\midrule
SARe-Gen
& 81.34 & 1.85 & 34.35 & 6.83 \\

SARe-Refine
& \textbf{84.89}
& \textbf{1.72}
& \textbf{32.51}
& \textbf{6.19} \\
\midrule
$\Delta$
& +3.55 & -0.13 & -1.84 & -0.64 \\
\bottomrule
\end{tabular}
\caption{
Effect of SARe-Refine on difficult Breaking Bad cases.
Results are averaged over the $N=2411$ pooled Breaking Bad test samples whose first-pass PA is below 95\%.
}
\label{tab:refine_hard}
\end{table}

\subsection{Qualitative Evaluations}

\begin{figure*}[t]
  \centering
  \includegraphics[width=0.99\textwidth]{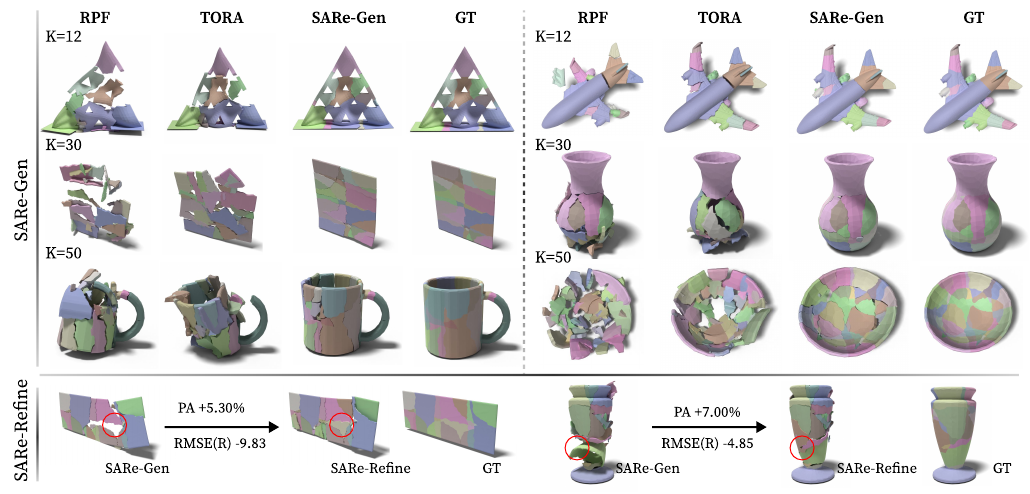}
    \caption{Qualitative comparisons of SARe-Gen and SARe-Refine. Red circles highlight regions corrected by the second pass.}
  \label{fig:gallery}
\end{figure*}

Fig.~\ref{fig:gallery} compares SARe-Gen with two baselines across fragment counts.
While all methods produce plausible results for small $K$, the baselines accumulate incorrect fragment contacts as $K$ increases, distorting global shapes, whereas SARe-Gen better preserves structural consistency.
The bottom row further shows that SARe-Refine corrects local assembly errors through verified-subassembly-guided resampling while preserving consistent regions.

\begin{figure}[t]
  \centering
  \includegraphics[width=0.99\columnwidth]{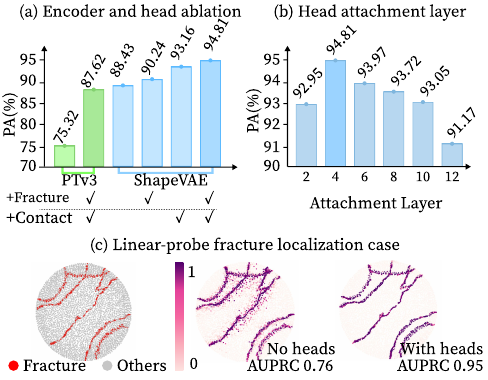}
  \caption{Ablations of model components and head attachment layers, with a fracture-localization probe.}
  \label{fig:ablation}
\end{figure}

\subsection{Ablation Studies and Analyses}
Unless otherwise stated, all analyses in this subsection use the Breaking Bad Everyday subset.
\noindent{\textbf{Component and Attachment-Layer Ablations.}}
Figure~\ref{fig:ablation} analyzes the geometry encoder, structural heads, and their attachment layer.
As shown in Fig.~\ref{fig:ablation}(a), adding the fracture-region and contact-graph heads improves PA from 75.32\% to 87.62\% with PTv3, showing that structural supervision is effective beyond the ShapeVAE encoder.
With ShapeVAE, the fracture-region and contact-graph heads individually improve PA from 88.43\% to 90.24\% and 93.16\%, respectively, while combining both achieves the best result of 94.81\%.
Figure~\ref{fig:ablation}(b) shows that attaching the heads to the intermediate layer 4 yields the highest PA, whereas very early or late attachment performs worse.
Figure~\ref{fig:ablation}(c) applies the linear fracture probe to frozen layer-4 tokens on a representative held-out bowl assembly ($K=12$).
Without structural-head training, the probe produces diffuse responses over broad non-fracture regions, whereas structural-head training concentrates predictions along fracture boundaries and suppresses off-fracture activations, improving the case-level AUPRC from 0.76 to 0.95.
Further probe details are provided in the supplementary.

\noindent{\textbf{Test-Time Query Scaling and Allocation.}}
Table~\ref{tab:testtime_scaling} evaluates test-time query-budget scaling and allocation using the same checkpoint trained with $M=5120$, without retraining.
Increasing $M$ improves performance primarily in the many-fragment regime, raising PA from 83.05\% to 84.89\% for $K=21$--$50$, at the cost of additional inference time.
Under the same total budget, uniform allocation substantially degrades PA, especially for $K=21$--$50$, demonstrating the importance of area-adaptive query allocation.
\begin{table}[t]
\centering
\setlength{\tabcolsep}{3.2pt}

\begin{tabular}{cccc}
\toprule
Setting
& \makecell{Time/obj.\\(s)}
& \makecell{PA$_{\mathrm{all}}$$\uparrow$\\(\%)}
& \makecell{PA$_{K=21\text{--}50}$$\uparrow$\\(\%)} \\
\midrule
$M=2560$  & 0.79 & 93.06 &  69.35 \\
$M=5120$  & 1.03 & 94.81 &  83.05 \\
$M=10240$ & 3.13 & \textbf{95.19} &  \textbf{84.89} \\
\midrule
Uniform ($M=5120$) & 1.02 & 79.69 &  44.96 \\
\bottomrule
\end{tabular}
\caption{
Test-time query-budget scaling and query-allocation ablation using the same checkpoint trained with $M=5120$.
}
\label{tab:testtime_scaling}
\end{table}

\section{Conclusion}
We present SARe, a structure-aware framework for many-fragment 3D reassembly.
SARe-Gen combines query-aligned geometry with fracture-region and fragment-contact supervision, while SARe-Refine verifies predicted contacts and uses reliable subassemblies to guide second-pass resampling.
Experiments across synthetic, scan-derived, and physically fractured objects demonstrate strong generalization and improved robustness as fragment count increases.
Future work will explore hierarchical assembly and appearance cues for highly ambiguous or incomplete inputs.

\bibliography{arxiv}

@String(CVPR  = {IEEE Conf. Comput. Vis. Pattern Recog.})

@String(ICCV  = {Int. Conf. Comput. Vis.})

@String(ECCV  = {Eur. Conf. Comput. Vis.})

@String(NeurIPS = {Adv. Neural Inform. Process. Syst.})

@String(ICML  = {Int. Conf. Mach. Learn.})

@String(ICLR  = {Int. Conf. Learn. Represent.})

@String(TOG   = {ACM Trans. Graph.})

@String(CVPR  = {CVPR})

@String(ICCV  = {ICCV})

@String(ECCV  = {ECCV})

@String(NeurIPS = {NeurIPS})

@String(ICML  = {ICML})

@String(ICLR  = {ICLR})

@String(TOG   = {ACM TOG})

@article{heritage1,
author = {Calzado-Mart\'{\i}nez, Alberto and Garc\'{\i}a-Fern\'{a}ndez, \'{A}ngel-Luis and Ortega-Alvarado, Lidia M. and Feito-Higueruela, Francisco-Ram\'{o}n},
title = {Integrated Information System for 3D Interactive Reconstruction of an Archaeological Site},
year = {2023},
issue_date = {September 2023},
publisher = {Association for Computing Machinery},
address = {New York, NY, USA},
volume = {16},
number = {3},
issn = {1556-4673},
doi = {10.1145/3586077},
journal = {J. Comput. Cult. Herit.},
articleno = {44},
numpages = {23}
}

@Article{accident1,
AUTHOR = {Jia, Caiqin and Ren, Yali and Wang, Zhi and Zhang, Yuan},
TITLE = {GeoAssemble: A Geometry-Aware Hierarchical Method for Point Cloud-Based Multi-Fragment Assembly},
JOURNAL = {Sensors},
VOLUME = {25},
YEAR = {2025},
NUMBER = {21},
URL = {https://www.mdpi.com/1424-8220/25/21/6533},
DOI = {10.3390/s25216533}
}

@inproceedings{robotic1,
  title={BiAssemble: Learning Collaborative Affordance for Bimanual Geometric Assembly},
  author={Yan Shen and Ruihai Wu and Yubin Ke and Xinyuan Song and Zeyi Li and Xiaoqi Li and Hongwei Fan and Haoran Lu and Hao Dong},
  booktitle=ICML,
  year={2025},
}

@inproceedings{heritage2-RePAIR, 
title = {The RePAIR Project: Datasets for archaeological and restoration studies in Pompeii}, 
booktitle = {Digital Heritage}, 
author = {Zuchtriegel, Gabriel and Brunetto, Maria Antonella and Gravina, Elena and Napolitano, Maria Cristina and Ricciardi, Francesca Simona and Zambrano, Alessandra and Khoroshiltseva, Marina and Palmieri, Luca and Pelillo, Marcello and Vascon, Sebastiano and Elkin, Gur and Shahar, Ofir Itzhak and Ohayon, Yaniv and Alali, Nadav and Ben-Shahar, Ohad}, 
isbn = {978-3-03868-277-6}, 
publisher = {The Eurographics Association}, 
doi = {10.2312/dh.20253389}, 
year = {2025},
}

@InProceedings{robotic2,
title = 	 {Blocks Assemble! Learning to Assemble with Large-Scale Structured Reinforcement Learning},
author =       {Ghasemipour, Seyed Kamyar Seyed and Kataoka, Satoshi and David, Byron and Freeman, Daniel and Gu, Shixiang Shane and Mordatch, Igor},
booktitle = 	 {Proceedings of the 39th International Conference on Machine Learning},
pages = 	 {7435--7469},
year = 	 {2022},
editor = 	 {Chaudhuri, Kamalika and Jegelka, Stefanie and Song, Le and Szepesvari, Csaba and Niu, Gang and Sabato, Sivan},
volume = 	 {162},
series = 	 {Proceedings of Machine Learning Research},
publisher =    {PMLR},
}

@inproceedings{Tsesmelis2024RePAIR,
 author = {Tsesmelis, Theodore and Palmieri, Luca and Khoroshiltseva, Marina and Islam, Adeela and Elkin, Gur and Shahar, Ofir Itzhak and Scarpellini, Gianluca and Fiorini, Stefano and Ohayon, Yaniv and Alali, Nadav and Aslan, Sinem and Morerio, Pietro and Vascon, Sebastiano and Gravina, Elena and Napolitano, Maria Cristina and Scarpati, Giuseppe and Zuchtriegel, Gabriel and Sp\"{u}hler, Alexandra and Fuchs, Michel E. and James, Stuart and Ben-Shahar, Ohad and Pelillo, Marcello and Del Bue, Alessio},
 booktitle = NeurIPS,
 doi = {10.52202/079017-0947},
 editor = {A. Globerson and L. Mackey and D. Belgrave and A. Fan and U. Paquet and J. Tomczak and C. Zhang},
 pages = {30076--30105},
 title = {Re-assembling the past: The RePAIR dataset and benchmark for real world 2D and 3D puzzle solving},
 volume = {37},
 year = {2024}
}

@article{Lu2025SurveyAssembly,
  title={A Survey on Computational Solutions for Reconstructing Complete Objects by Reassembling Their Fractured Parts},
  author={Jiaxin Lu and Yongqing Liang and Huijun Han and Jiacheng Hua and Junfeng Jiang and Xin Li and Qi-Xing Huang},
  journal={Computer Graphics Forum},
  year={2024},
  volume={44},
}

@article{lee2026tora,
  title     = {TORA: Topological Representation Alignment for 3D Shape Assembly},
  author    = {Lee, Nahyuk and Chen, Zhiang and Pollefeys, Marc and Hong, Sunghwan},
  journal   = {arXiv preprint arXiv:2604.04050},
  year      = {2026}
}

@article{Funkhouser2011,
title        = {Learning How to Match Fresco Fragments},
author       = {Funkhouser, Thomas and Shin, Hijung and Toler-Franklin, Corey and Garc{\'\i}a Casta{\~n}eda, Antonio and Brown, Benedict and Dobkin, David and Rusinkiewicz, Szymon and Weyrich, Tim},
journal      = {ACM Journal on Computing and Cultural Heritage},
volume       = {4},
number       = {2},
year         = {2011},
publisher    = {Association for Computing Machinery},
address      = {New York, NY, USA},
doi          = {10.1145/2037820.2037824},
}

@inproceedings{Structure_From_Sherds,
  author={Hong, Je Hyeong and Yoo, Seong Jong and Zeeshan, Muhammad Arshad and Kim, Young Min and Kim, Jinwook},
  booktitle=ICCV, 
  title={Structure-from-Sherds: Incremental 3D Reassembly of Axially Symmetric Pots from Unordered and Mixed Fragment Collections}, 
  year={2021},
  pages={5423-5431},
  doi={10.1109/ICCV48922.2021.00539}}

@article{Huang2006Reassembling,
author = {Huang, Qi-Xing and Fl\"{o}ry, Simon and Gelfand, Natasha and Hofer, Michael and Pottmann, Helmut},
title = {Reassembling fractured objects by geometric matching},
year = {2006},
publisher = {Association for Computing Machinery},
address = {New York, NY, USA},
volume = {25},
number = {3},
doi = {10.1145/1141911.1141925},
journal = TOG,
pages = {569–578},
numpages = {10},
}

@article{Xu2015RobustSurface,
title = {Robust surface segmentation and edge feature lines extraction from fractured fragments of relics},
journal = {Journal of Computational Design and Engineering},
volume = {2},
number = {2},
pages = {79-87},
year = {2015},
issn = {2288-4300},
doi = {https://doi.org/10.1016/j.jcde.2014.12.002},
author = {Jiangyong Xu and Mingquan Zhou and Zhongke Wu and Wuyang Shui and Sajid Ali},
}

@inproceedings{Castaneda2011GlobalConsistency,
author = {Casta\~{n}eda, A. Garc\'{\i}a and Brown, B. and Rusinkiewicz, S. and Funkhouser, T. and Weyrich, T.},
title = {Global consistency in the automatic assembly of fragmented artefacts},
year = {2011},
isbn = {9783905674347},
publisher = {Eurographics Association},
booktitle = {Proceedings of the 12th International Conference on Virtual Reality, Archaeology and Cultural Heritage},
pages = {73–80},
numpages = {8},
location = {Prato, Italy},
}

@inproceedings{RPF,
      author = {Sun, Tao and Zhu, Liyuan and Huang, Shengyu and Song, Shuran and Armeni, Iro},
      title = {Rectified Point Flow: Generic Point Cloud Pose Estimation},
      booktitle = NeurIPS,
      year = {2025},
}

@inproceedings{GARF,
  author    = {Li, Sihang and Jiang, Zeyu and Chen, Grace and Xu, Chenyang and Tan, Siqi and Wang, Xue and Fang, Irving and Zyskowski, Kristof and McPherron, Shannon P. and Iovita, Radu and Feng, Chen and Zhang, Jing},
  title     = {GARF: Learning Generalizable 3D Reassembly for Real-World Fractures},
  booktitle = ICCV,
  month     = {October},
  year      = {2025},
  pages     = {5711--5721}
}

@inproceedings{PuzzleFusionPP,
  title     = {PuzzleFusion++: Auto-agglomerative 3D Fracture Assembly by Denoise and Verify},
  author    = {Wang, Zhengqing and Chen, Jiacheng and Furukawa, Yasutaka},
  booktitle = ICLR,
  year      = {2025},
}

@InProceedings{JigsawPP,
  author    = {Lu, Jiaxin and Hua, Gang and Huang, Qixing},
  title     = {Jigsaw++: Imagining Complete Shape Priors for Object Reassembly},
  booktitle = ICCV,
  month     = {October},
  year      = {2025},
  pages     = {6704--6714}
}

@inproceedings{FragmentDiff,
author = {Xu, Qun-Ce and Chen, Hao-Xiang and Hua, Jiacheng and Zhan, Xiaohua and Yang, Yong-Liang and Mu, Tai-Jiang},
title = {FragmentDiff: A Diffusion Model for Fractured Object Assembly},
year = {2024},
publisher = {Association for Computing Machinery},
address = {New York, NY, USA},
doi = {10.1145/3680528.3687673},
booktitle = {SIGGRAPH Asia 2024 Conference Papers},
numpages = {12},
}

@InProceedings{Scarpellini2024DiffAssemble,
    author    = {Gianluca Scarpellini and Stefano Fiorini and Francesco Giuliari and Pietro Morerio and Alessio {Del Bue}},
    title     = {DiffAssemble: A Unified Graph-Diffusion Model for 2D and 3D Reassembly},
    booktitle = CVPR,
    month     = {June},
    year      = {2024},
}

@InProceedings{Wu_2023_SE3,
  author={Wu, Ruihai and Tie, Chenrui and Du, Yushi and Zhao, Yan and Dong, Hao},
  booktitle=ICCV, 
  title={Leveraging SE(3) Equivariance for Learning 3D Geometric Shape Assembly}, 
  year={2023},
  pages={14265-14274},
  doi={10.1109/ICCV51070.2023.01316}
}

@inproceedings{BreakingBad,
 author = {Sell\'{a}n, Silvia and Chen, Yun-Chun and Wu, Ziyi and Garg, Animesh and Jacobson, Alec},
 booktitle = NeurIPS,
 pages = {38885--38898},
 title = {Breaking Bad: A Dataset for Geometric Fracture and Reassembly},
 volume = {35},
 year = {2022}
}

@inproceedings{Lu2023Jigsaw,
 author = {Lu, Jiaxin and Sun, Yifan and Huang, Qixing},
 booktitle = NeurIPS,
 pages = {14969--14986},
 publisher = {Curran Associates, Inc.},
 title = {Jigsaw: Learning to Assemble Multiple Fractured Objects},
 volume = {36},
 year = {2023}
}

@InProceedings{Qin2022Geometric,
  author={Qin, Zheng and Yu, Hao and Wang, Changiian and Guo, Yulan and Peng, Yuxing and Xu, Kai},
  booktitle=CVPR, 
  title={Geometric Transformer for Fast and Robust Point Cloud Registration}, 
  year={2022},
  pages={11133-11142},
  doi={10.1109/CVPR52688.2022.01086}
}

@inproceedings{Huang2020GPA,
    author = {Huang, Jialei and Zhan, Guanqi and Fan, Qingnan and Mo, Kaichun and Shao, Lin and Chen, Baoquan and Guibas, Leonidas and Dong, Hao},
    title = {Generative 3D Part Assembly via Dynamic Graph Learning},
    booktitle = NeurIPS,
    year = {2020}
}

@inproceedings{Zhao2025Assembler,
author = {Zhao, Wang and Cao, Yan-Pei and Xu, Jiale and Dong, Yuejiang and Shan, Ying},
title = {Assembler: Scalable 3D Part Assembly via Anchor Point Diffusion},
year = {2025},
publisher = {Association for Computing Machinery},
doi = {10.1145/3757377.3763972},
booktitle = {Proceedings of the SIGGRAPH Asia 2025 Conference Papers},
numpages = {11},
series = {SA Conference Papers '25}
}

@inproceedings{Lan2025GaussianAnything,
  title={GaussianAnything: Interactive Point Cloud Latent Diffusion for 3D Generation},
  author={Lan, Yushi and Zhou, Shangchen and Lyu, Zhaoyang and Hong, Fangzhou and Yang, Shuai and Dai, Bo and Pan, Xingang and Loy, Chen Change},
  year={2025},
  booktitle={ICLR},
}

@InProceedings{Lugmayr2022RePaint,
  author    = {Lugmayr, Andreas and Danelljan, Martin and Romero, Andres and Yu, Fisher and Timofte, Radu and Van Gool, Luc},
  title     = {RePaint: Inpainting Using Denoising Diffusion Probabilistic Models},
  booktitle = CVPR,
  month     = {June},
  year      = {2022},
  pages     = {11461--11471}
}

@InProceedings{Chen2022NeuralShape,
  author    = {Chen, Yun-Chun and Li, Haoda and Turpin, Dylan and Jacobson, Alec and Garg, Animesh},
  title     = {Neural Shape Mating: Self-Supervised Object Assembly With Adversarial Shape Priors},
  booktitle = CVPR,
  month     = {June},
  year      = {2022},
  pages     = {12724--12733}
}

@misc{Li2024GPAT,
  author = {Li, Jiahan and Cheng, Chaoran and Ma, Jianzhu and Liu, Ge},
  title  = {Geometric Point Attention Transformer for 3D Shape Reassembly},
  note   = {arXiv:2411.17788},
  year   = {2024}
}

@misc{Zhao2025Hunyuan3D2,
  author = {Hunyuan3d},
  title  = {Hunyuan3D 2.0: Scaling Diffusion Models for High Resolution Textured 3D Assets Generation},
  note   = {arXiv:2501.12202},
  year   = {2025}
}

@article{Zhang2023_3DShape2VecSet,
author = {Zhang, Biao and Tang, Jiapeng and Nie\ss{}ner, Matthias and Wonka, Peter},
title = {3DShape2VecSet: A 3D Shape Representation for Neural Fields and Generative Diffusion Models},
year = {2023},
issue_date = {August 2023},
publisher = {Association for Computing Machinery},
address = {New York, NY, USA},
volume = {42},
number = {4},
issn = {0730-0301},
doi = {10.1145/3592442},
journal = TOG,
}

@InProceedings{Ren2024Xcube,
  author={Ren, Xuanchi and Huang, Jiahui and Zeng, Xiaohui and Museth, Ken and Fidler, Sanja and Williams, Francis},
  booktitle = CVPR, 
  title={XCube: Large-Scale 3D Generative Modeling using Sparse Voxel Hierarchies}, 
  year={2024},
  pages={4209-4219},
  doi={10.1109/CVPR52733.2024.00403}
}

@inproceedings{Zeng2022LION,
 author = {Zeng, Xiaohui and Vahdat, Arash and Williams, Francis and Gojcic, Zan and Litany, Or and Fidler, Sanja and Kreis, Karsten},
 booktitle = NeurIPS,
 editor = {S. Koyejo and S. Mohamed and A. Agarwal and D. Belgrave and K. Cho and A. Oh},
 pages = {10021--10039},
 publisher = {Curran Associates, Inc.},
 title = {LION: Latent Point Diffusion Models for 3D Shape Generation},
 url = {https://proceedings.neurips.cc/paper_files/paper/2022/file/40e56dabe12095a5fc44a6e4c3835948-Paper-Conference.pdf},
 volume = {35},
 year={2022}
}

@InProceedings{TRELLIS,
    title   = {Structured 3D Latents for Scalable and Versatile 3D Generation},
    author  = {Xiang, Jianfeng and Lv, Zelong and Xu, Sicheng and Deng, Yu and Wang, Ruicheng and 
               Zhang, Bowen and Chen, Dong and Tong, Xin and Yang, Jiaolong},
  booktitle = CVPR,
  month     = {June},
  year      = {2025},
  pages     = {21469--21480}
}

@inproceedings{Li2025Sparc3D,
  author    = {Li, Zhihao and Wang, Yufei and Zheng, Heliang and Luo, Yihao and Wen, Bihan},
  title     = {Sparc3D: Sparse Representation and Construction for High-Resolution 3D Shapes Modeling},
  booktitle = NeurIPS,
  year      = {2025},
}

@inproceedings{zhang2025lagem,
  title     = {{LaGeM}: A Large Geometry Model for 3D Representation Learning and Diffusion},
  author    = {Biao Zhang and Peter Wonka},
  booktitle = ICLR,
  year      = {2025},
  url       = {https://openreview.net/forum?id=72OSO38a2z}
}

@InProceedings{Chen2025Dora,
    author    = {Chen, Rui and Zhang, Jianfeng and Liang, Yixun and Luo, Guan and Li, Weiyu and Liu, Jiarui and Li, Xiu and Long, Xiaoxiao and Feng, Jiashi and Tan, Ping},
    title     = {Dora: Sampling and Benchmarking for 3D Shape Variational Auto-Encoders},
    booktitle = CVPR,
    month     = {June},
    year      = {2025},
    pages     = {16251-16261}
}

@inproceedings{Chen2025AutoPartGen,
  title     = {AutoPartGen: Autoregressive 3D Part Generation and Discovery},
  author    = {Chen, Minghao and Wang, Jianyuan and Shapovalov, Roman and Monnier, Tom and Jung, Hyunyoung and Wang, Dilin and Ranjan, Rakesh and Laina, Iro and Vedaldi, Andrea},
  booktitle = NeurIPS,
  year      = {2025},
  url       = {https://openreview.net/forum?id=ljJGBcpn7q},
}

@inproceedings{Pernias2024Wuerstchen,
title     = {W{\"u}rstchen: An Efficient Architecture for Large-Scale Text-to-Image Diffusion Models},
author    = {Perni{\'a}s, Pablo and Rampas, Dominic and Richter, Mats L. and Pal, Christopher J. and Aubreville, Marc},
booktitle = {International Conference on Learning Representations (ICLR)},
year      = {2024},
url       = {[https://openreview.net/forum?id=gU58d5QeGv](https://openreview.net/forum?id=gU58d5QeGv)},
}

@inproceedings{Li2024ReturnOfUnconditionalGeneration,
 author = {Li, Tianhong and Katabi, Dina and He, Kaiming},
 booktitle = NeurIPS,
 doi = {10.52202/079017-3985},
 pages = {125441--125468},
 publisher = {Curran Associates, Inc.},
 title = {Return of Unconditional Generation: A Self-supervised Representation Generation Method},
 volume = {37},
 year = {2024}
}

@inproceedings{yu2025repa,
title     = {Representation Alignment for Generation: Training Diffusion Transformers Is Easier Than You Think},
author    = {Yu, Sihyun and Kwak, Sangkyung and Jang, Huiwon and Jeong, Jongheon and Huang, Jonathan and Shin, Jinwoo and Xie, Saining},
booktitle = ICLR,
year      = {2025},
url       = {[https://openreview.net/forum?id=DJSZGGZYVi](https://openreview.net/forum?id=DJSZGGZYVi)},
}

@InProceedings{Guo_2023_CVPR,
  title={Shadowdiffusion: When degradation prior meets diffusion model for shadow removal},
  author={Guo, Lanqing and Wang, Chong and Yang, Wenhan and Huang, Siyu and Wang, Yufei and Pfister, Hanspeter and Wen, Bihan},
  booktitle=CVPR,
  pages={14049--14058},
  year={2023}
}

@misc{Xu2025TopoDiffuser,
author = {Xu, Zehui and Wang, Junhui and Shi, Yongliang and Gao, Chao and Zhou, Guyue},
title  = {TopoDiffuser: A Diffusion-Based Multimodal Trajectory Prediction Model with Topometric Maps},
note   = {arXiv:2508.00303},
year   = {2025}
}

@InProceedings{Ye2024DiffusionMTL,
  title={DiffusionMTL: Learning Multi-Task Denoising Diffusion Model from Partially Annotated Data},
  author={Ye, Hanrong and Xu, Dan},
  booktitle={CVPR},
  year={2024}
}

@inproceedings{Yang2025TaskDiffusion,
title     = {Multi-Task Dense Predictions via Unleashing the Power of Diffusion},
author    = {Yang, Yuqi and Jiang, Peng-Tao and Hou, Qibin and Zhang, Hao and Chen, Jinwei and Li, Bo},
booktitle = ICLR,
year      = {2025},
}

@inproceedings{Liu2023RectifiedFlow,
  title     = {Flow Straight and Fast: Learning to Generate and Transfer Data with Rectified Flow},
  author    = {Liu, Xingchao and Gong, Chengyue and Liu, Qiang},
  booktitle = ICLR,
  year      = {2023},
  url       = {https://openreview.net/forum?id=XVjTT1nw5z},
}

@misc{Liu2022RectifiedFlowOT,
  author = {Liu, Qiang},
  title  = {Rectified Flow: A Marginal Preserving Approach to Optimal Transport},
  note   = {arXiv:2209.14577},
  year   = {2022}
}

@inproceedings{Tancik2020FourierFeatures,
  title     = {Fourier Features Let Networks Learn High Frequency Functions in Low Dimensional Domains},
  author    = {Tancik, Matthew and Srinivasan, Pratul P. and Mildenhall, Ben and Fridovich-Keil, Sara and Raghavan, Nithin and Singhal, Utkarsh and Ramamoorthi, Ravi and Barron, Jonathan T. and Ng, Ren},
  booktitle = NeurIPS,
  volume    = {33},
  year      = {2020},

}

@inproceedings{Mildenhall2020NeRF,
  title        = {NeRF: Representing Scenes as Neural Radiance Fields for View Synthesis},
  author       = {Ben Mildenhall and
                  Pratul P. Srinivasan and
                  Matthew Tancik and
                  Jonathan T. Barron and
                  Ravi Ramamoorthi and
                  Ren Ng},
  booktitle = ECCV,
  year      = {2020},
  publisher = {Springer International Publishing},
  doi       = {10.1007/978-3-030-58452-8_24},
  url       = {https://doi.org/10.1007/978-3-030-58452-8_24}
}

@InProceedings{Go2025SplatFlow,
  author    = {Go, Hyojun and Park, Byeongjun and Jang, Jiho and Kim, Jin-Young and Kwon, Soonwoo and Kim, Changick},
  title     = {SplatFlow: Multi-View Rectified Flow Model for 3D Gaussian Splatting Synthesis},
  booktitle = CVPR,
  month     = {June},
  year      = {2025},
  pages     = {21524--21536}
}

@inproceedings{wu2023omniobject3d,
  author={Wu, Tong and Zhang, Jiarui and Fu, Xiao and Wang, Yuxin and Ren, Jiawei and Pan, Liang and Wu, Wayne and Yang, Lei and Wang, Jiaqi and Qian, Chen and Lin, Dahua and Liu, Ziwei},
  booktitle=CVPR, 
  title={OmniObject3D: Large-Vocabulary 3D Object Dataset for Realistic Perception, Reconstruction and Generation}, 
  year={2023},
  volume={},
  number={},
  pages={803-814},
  doi={10.1109/CVPR52729.2023.00084}
}

@inproceedings{lamb2023fantastic,
  author={Lamb, Nikolas and Palmer, Cameron and Molloy, Benjamin and Banerjee, Sean and Banerjee, Natasha Kholgade},
  booktitle=CVPR, 
  title={Fantastic Breaks: A Dataset of Paired 3D Scans of Real-World Broken Objects and Their Complete Counterparts}, 
  year={2023},
  pages={4681-4691},
  doi={10.1109/CVPR52729.2023.00454}
}

@article{Peebles2022DiT,
  title={Scalable Diffusion Models with Transformers},
  author={William Peebles and Saining Xie},
  year={2022},
  journal={arXiv preprint arXiv:2212.09748},
}

@article{breakinggood,
author = {Sell\'{a}n, Silvia and Luong, Jack and Mattos Da Silva, Leticia and Ramakrishnan, Aravind and Yang, Yuchuan and Jacobson, Alec},
title = {Breaking Good: Fracture Modes for Realtime Destruction},
year = {2023},
publisher = {Association for Computing Machinery},
address = {New York, NY, USA},
volume = {42},
number = {1},
url = {https://doi.org/10.1145/3549540},
doi = {10.1145/3549540},
journal = ToG,
numpages = {12},
}


\end{document}